%% file: main.tex
\documentclass[sigconf]{acmart}
\AtBeginDocument{%
  \providecommand\BibTeX{{%
    \normalfont B\kern-0.5em{\scshape i\kern-0.25em b}\kern-0.8em\TeX}}}
    
\renewcommand\footnotetextcopyrightpermission[1]{}
\setcopyright{none}
\usepackage[font=small,labelfont=bf,skip=0pt]{caption}
\usepackage{multirow}
\usepackage{tabularx}
\usepackage{mathtools}
\usepackage{balance}
\usepackage{bbm}
\usepackage{color}
\usepackage{float}
\usepackage{listings}
\usepackage{color}
\usepackage{booktabs}
\usepackage{tablefootnote}
\usepackage{url}
\usepackage{threeparttable}
\usepackage{subcaption}
\usepackage{titlesec}
\usepackage{comment}
\usepackage{stfloats}
\usepackage{fancyhdr}

\titlespacing\section{0pt}{7pt plus 0pt minus 2pt}{0pt plus 0pt minus 2pt}
\titlespacing\subsection{0pt}{6pt plus 0pt minus 2pt}{0pt plus 0pt minus 2pt}

\newcommand\possiblebreak{\ifhmode\unskip\space\hfil\penalty0\hfilneg\fi}

\usepackage{adjustbox}
\usepackage{algorithmic}
\usepackage[ruled,vlined]{algorithm2e}
\SetAlFnt{\small}

\title{Zhuyi: Perception Processing Rate Estimation\\for Safety in Autonomous Vehicles}

\author{\normalsize Yu-Shun Hsiao$^\ddagger$, Siva Kumar Sastry Hari$^\dagger$, Micha\l~Filipiuk$^\dagger$, Timothy Tsai$^\dagger$, Michael B. Sullivan$^\dagger$, \\ Vijay Janapa Reddi$^\ddagger$, Vasu Singh$^\dagger$, and Stephen W. Keckler$^\dagger$ \\ $^\ddagger$ Harvard University, $^\dagger$ NVIDIA Corporation}

\begin{document}
\sloppy
\begin{abstract}
The processing requirement of autonomous vehicles (AVs) for high-accuracy perception in complex scenarios can exceed the resources offered by the in-vehicle computer, degrading safety and comfort. This paper proposes a sensor frame processing rate (FPR) estimation model, Zhuyi, that quantifies the minimum safe FPR continuously in a driving scenario. Zhuyi can be employed post-deployment as an online safety check and to prioritize work. Experiments conducted using a multi-camera state-of-the-art industry AV system show that Zhuyi's estimated FPRs are conservative, yet the system can maintain safety by processing only 36\% or fewer frames compared to a default 30-FPR system in the tested scenarios. 
\end{abstract}


\settopmatter{printfolios=true}
\pagestyle{plain} 
\maketitle

\input{1_Introduction}
\input{3_Methodology}
\input{4_System}
\input{5_Experiments}
\input{2_Related}
\input{6_Conclusion}

\bibliographystyle{abbrv}
\footnotesize{%
\bibliography{main}
}
\end{document}

%% file: 1_Introduction.tex
\section{INTRODUCTION}
\label{intro}
Safety remains an important consideration as autonomous vehicles (AVs) become increasingly complex. AV systems are equipped with a growing number of sensors to improve perception of the surrounding environment, which also generates more work for the perception processing. 
AVs are expected to employ about a dozen high-resolution cameras, along with multiple radars, LiDARs, and other sensors (e.g., ultra-sonic sensors and high-precision positioning units)~\cite{Hyperion8, NIO_NAD}. 
Data from the sensors is processed using multiple perception models to extract information that enables a safe and comfortable drive. Perception models include object detection/segmentation, lane detection, free space perception, signal light perception, road sign recognition, sensor occlusion detection, and hazard detection~\cite{drive}. In addition to perception, complex algorithms can predict trajectories, plan a route, test for safety, and control motion. Driver assistance software will also be included in future AVs. All of the computation demand is serviced by an in-vehicle computer. 

\begin{figure}[t]
    \centering
    \includegraphics[width=0.9\columnwidth]{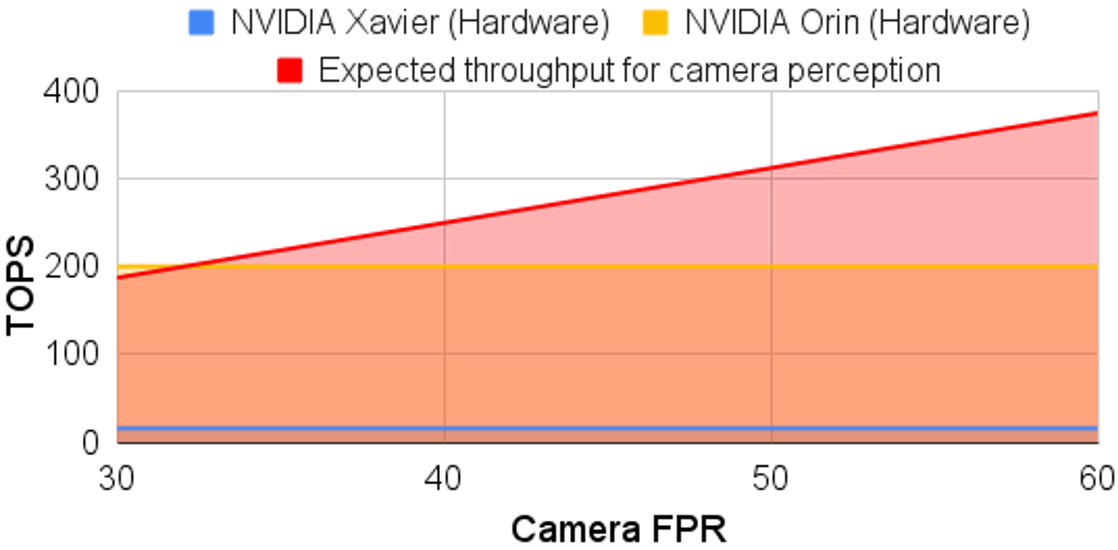}
    \caption{Expected throughput demand for state-of-the-art camera perception with MLPerf models~\cite{reddi2020mlperf}.}
    \label{fig:motivation}
    \vspace{-5pt}
\end{figure}

A preliminary investigation suggests that the processing power for high-quality perception alone on a 12-camera system can exceed the resources offered by a state-of-the-art SoC. Figure~\ref{fig:motivation} shows the throughput demand of running the perception tasks using state-of-the-art models along with the throughput offered by single NVIDIA DRIVE AGX Xavier and NVIDIA Jetson AGX Orin SoCs~\cite{Orin}. We estimate the 
Tera Operations Per Second (TOPS)
assuming the SSD-Large object detection model is run for 1200x1200 pixel frames on all 12 cameras (requirement per run is from MLPerf~\cite{reddi2020mlperf}). Since accurate perception also requires running other camera-based models (e.g., lane detection, free space perception, occlusion detection), we increase the demand by 20\%, assuming the additional models can reuse the extracted features. The compute requirements could grow even higher by considering LiDARs, radars, and localization and planning algorithms. The perception algorithms could require multiple times more computation with higher input resolution. 

Moreover, the work may increase based on the scenarios. For example, a recent study showed that higher obstacle density around an AV can intermittently increase the computational demand~\cite{zhao2020driving}. As the compute demand created by real-time AV systems can be very high, it is important to quantify the perception requirement for safe operation, provision the fixed resources in the in-vehicle computer to prioritize important tasks, and use the leftover resources for tasks that will further improve safety and comfort.

This paper proposes a processing rate estimation model, Zhuyi, to {\it quantify the tolerable per-camera processing latency in a multi-camera setting for safe operation.} 
We define safe operation or safety as no collision between the ego and surrounding actors\footnote{The AV is referred to as the \textit{ego}. Dynamic objects in a scenario are \textit{actors}. \\\\© 2022 IEEE. Personal use of this material is permitted. Permission from IEEE must be obtained for all other uses, in any current or future media, including reprinting/republishing this material for advertising or promotional purposes, creating new collective works, for resale or redistribution to servers or lists, or reuse of any copyrighted component of this work in other works.}. Based on the current and future states of the ego and surrounding actors, we apply a kinematics model that considers the possibility of a collision to quantify the maximum tolerable latency for safe driving. The reciprocal of the maximum tolerable latency is the minimum frame processing rate or FPR requirement for a camera sensor. Therefore, instead of a fixed sensor setting (e.g., 30 FPR per camera) in a typical AV system, the AV system can allocate hardware resources to process safety-critical perception tasks and avoid making an untimely decision or increasing reaction time.

Zhuyi has many use cases. During the development phase of the AV software system, Zhuyi can be used to verify whether the resource allocation for different AV tasks is sufficient for safety and provide input to design a more effective system. AV software is tested in a diverse set of driving scenarios for different operational design domains (ODDs)~\cite{odd2016} before deployment. 
For each driving scenario, Zhuyi can provide a maximum tolerable latency requirement per camera to avoid collisions. Zhuyi can also be employed as part of regression testing.
The analysis of latency estimates can be leveraged to better design the AV system. For instance, the latency boundary provided by Zhuyi can accelerate design space exploration and help architects to discover new optimization opportunities for different ODDs.

Zhuyi can also be used at runtime to improve the safety of the AV system. 
A Zhuyi-based system utilizes the ego's and actors' current and predicted future states for per-camera processing rate estimation. This estimate can be used as an online safety check or for workload prioritization. As a safety check, the processing latency for each camera is checked to be less than the maximum tolerable latency according to Zhuyi, thus ensuring that the reaction time of the AV system is fast enough to avoid any potential collisions. Zhuyi's estimate can also be used to prioritize work to ensure that the available hardware resources are utilized based on importance. For example, the processing rate for a less important camera that is sensing/tracking unimportant obstacles can be lowered when sufficient resources are not available to process data from a more important sensor, optimizing the AV system for comfort and safety. The dynamic FPR adjustment is especially critical when the hardware system is constrained due to operating conditions or increased delays for some tasks. Zhuyi facilitates hardware resource allocation for important computing tasks to enhance safety in safety-critical scenarios.

The contributions of our work are as follows: 
\begin{itemize}
    \item We propose Zhuyi to estimate each sensor's frame processing rate requirement that maintains safety.
    \item We present applications of Zhuyi, which includes a Zhuyi-based AV system that performs an online safety check and prioritizes work to improve safety by better managing available hardware resources.
    \item We demonstrate that with a Zhuyi-based AV system, only 36\% or fewer frames need to be processed for safety in the studied driving scenarios, assuming the baseline system is provisioned to process 30 camera frames per second. 
\end{itemize}

%% file: 3_Methodology.tex

\section{ZHUYI MODEL}
\label{sec:methodology} 
This section explains the Zhuyi model that estimates the per-camera frame processing rate (FPR) for a safe and comfortable drive.  The Zhuyi model employs kinematics-based calculations to estimate the processing latency requirement per actor and aggregates the requirements based on the cameras' field of view to obtain the FPR.

\subsection{Estimating Tolerable Latency Per Actor} 

\begin{figure}[t]
\centering
    \subfloat[Ego and actor locations at $t_0$ and $t_n$. \label{fig:location}]{\includegraphics[width=0.54\linewidth]{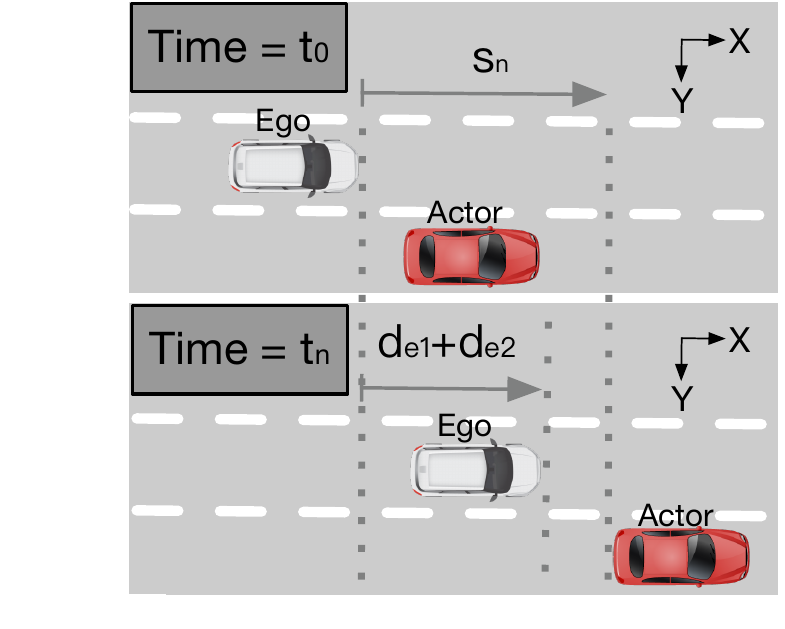}} 
    \subfloat[Ego velocity versus time graph. \label{fig:velocity}]{\includegraphics[width=0.46\linewidth]{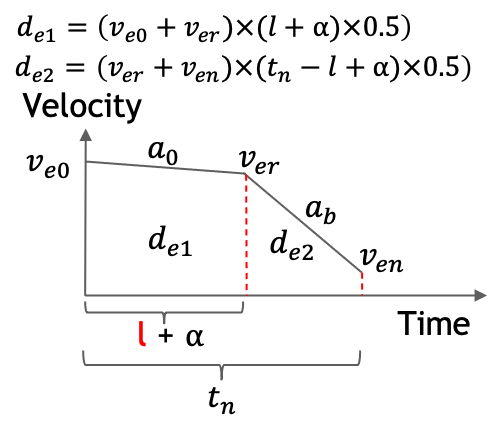}}
    \vspace{-2pt}
    \caption[]{Ego and actor states.}
    \vspace{-7pt}
    \label{fig:zhuyi-model}
\end{figure}

At a given time in a driving scenario, Zhuyi considers the current and future states of the ego and each surrounding actor individually to determine if a collision is possible. Figure~\ref{fig:location} shows the ego and the actor locations at times $t_0$ and $t_n$. We define the world reference frame as the 2-D top view with $X$ in the longitudinal direction of the ego and $Y$ in the lateral direction of the ego. The states at both $t_0$ and $t_n$ are shown in 2-D $XY$ frames in Figure~\ref{fig:zhuyi-model}. At $t_n$ (assuming $t_0=0$ and $t_n=n$  for simplicity), we desire no safety incident to take place, i.e., no collision between the ego and actor in either the longitudinal or lateral directions.

For the ego to respond to an obstacle, it must perceive the obstacle and react to it. We refer to the time used by the ego to perceive and confirm the obstacle as the reaction time ($t_r$) and the distance traveled during this time as $d_{e1}$.
After this period, the ego employs a safety procedure to avoid a potential collision. Assuming hard-braking as the safety procedure, the ego can travel $d_{e2}$ using a deceleration of $a_{b}$ reaching $v_{en}$ velocity. We use $a_b=max(C3, C4\times a_0)$, where C3 is the minimum braking deceleration and C4 is to account for the fact that the braking deceleration can be higher than the current value. 
For the ego to have no collision during $t_n$, the distance traveled by the ego must be less than the distance between the ego at time $t_0$ and actor at $t_n$ (we refer to this distance as $s_n$), and the velocity of the ego must be less than that of the actor. 
These two constraints are shown in Equations~\ref{eq:condition1} and ~\ref{eq:condition2}, respectively.

\vspace{-12pt}
\begin{equation}
\label{eq:condition1}
	d_{e1} + d_{e2} \leq s_n \times C1
	\vspace{-15pt}
\end{equation} 

\begin{equation}
\label{eq:condition2}
	v_{an} \times C2 \ge v_{en} \geq 0
	\vspace{-3pt}
\end{equation}
where $v_{an}$ is the velocity of the actor at time $t_n$ and $C1$ and $C2$ are constants used to add conservatism to the formulation.

The reaction time ($t_r$) is defined as $l + \alpha$, where $l$ is the tolerable latency and $\alpha$ accounts for the actor confirmation delay. 
In this paper, we model $\alpha$ as $K \times (l - l_0)$, where $K$ is the number of frames the perception system takes to confirm an actor and $l_0$ is the processing latency of the system at $t_0$. Based on the smoothing/filtering algorithm employed by the perception solution, a different model can be used to estimate actor confirmation delay. 
During $t_r$, we assume the ego's acceleration is unchanged.

\textit{Our objective is to find the maximum $latency$ that satisfies constraints in Equations~\ref{eq:condition1} and~\ref{eq:condition2} for any $t_n$ such that $t_n \ge t_r > t_0$.}
One algorithm to find the tolerable $latency$ is to iteratively decrease $l$ (from a maximum allowed value, e.g., 1s) and search for a $t_n$ where the two constraints  are met. We terminate the search as soon as the constraints are met. 
For an $l$, the algorithm sets $t'_n=l+\alpha$ (where $d_{e2}=0$) and checks whether the constraints are met using the ego and actor locations, velocities, and accelerations.
If the constraints are not met, a naive approach is to increment $t'_n$ by one timestep (e.g., 0.01s) and re-check. To improve performance, $\delta t_n$ can be computed based on the unmet constraint(s) to adjust $t'_n$ for the next iteration.
For the distance and velocity based constraints, this optimization computes how long the ego takes to cover the gap ($gap_d=s_n\times C1-d_{e1}-d{e2}$) and to reach the target velocity ($v_{an}\times C2-v_{en})$, respectively. Equation~\ref{eq:deltat} shows how $\delta t_n$ is computed. Using this $\delta t_n$, f $t'_n$ is updated ($t'_n=t'_n+\delta t_n$), and the constraint checking and $t'_n$ update steps are repeated $M$ times (e.g., $M=10$). If the constraints are not met after $M$ attempts, the procedure is repeated after decreasing $l$ by $\delta l$ (e.g., $\delta l=33ms$) until $l$ reaches the minimum allowed value (e.g., 33ms), resulting in a maximum of $L=max(l)/\delta l$ steps. 
\vspace{-5pt}
\begin{equation}
\label{eq:deltat}
    \delta t_n = 
    \begin{cases}
        \delta t_n^d = \frac{v_{en}+\sqrt{v_{en}^2 + 2 \times a_b \times |gap_d|}}{a_b},
        \\
        \qquad \qquad \qquad \text{iff } gap_d = C1 \times s_n - d_{e1} - d_{e2} \ge 0
        \\
        \delta t_n^v = \frac{gap_v}{a_b}, \quad \text{ iff } gap_v = v_{en} - C2 \times v_{an} \ge 0
        \\
        min(\delta t_n^d, \delta t_n^v), \; \text{otherwise}
    \end{cases}
\end{equation}

So far, we have estimated the tolerable latency for each actor assuming a single future trajectory. During operation, the AV may predict multiple future trajectories, each with an associated probability, for each actor in the scene.  Zhuyi estimates tolerable latency for all the predicted trajectories and considers different aggregation functions to obtain a single estimate per actor. For instance, maximum provides the most pessimistic estimate, and average gives more weight to the most likely future trajectory. $n_{th}$ percentile of the tolerable latency can also be used (e.g., $n=99$) as shown in Equation~\ref{eq:object}, which allows the ego to be cautious while being not too pessimistic.
\vspace{-5pt}
\begin{equation}
\label{eq:object}
	l_{\mathrm{actor}} = \mathrm{PR}_{n_{th}} \{l_j\} \mathrm{, for} j \in T
\end{equation} 
where $T$ is the set of predicted trajectories given by a trajectory predictor. Significant research has been conducted on trajectory prediction models~\cite{predictionnet, multipath, multiplefutures}, which we leverage in this work.

\subsection{Processing Rate Requirement Per Camera} 
Having the per-actor tolerable latency estimation model derived, Zhuyi obtains the per-camera FPR requirement by considering all the surrounding actors within each camera's field of view (FOV). 
Equation~\ref{eq:sensor} can be used to obtain the FPR per camera. As a result, Zhuyi can estimate per-camera FPR such that the AV can operate safely (collision-free) while using fewer computational resources. 

\vspace{-5pt}
\begin{align}
\label{eq:sensor}
	&\mathrm{FPR}_{\mathrm{sensor}} = \dfrac{1}{l_{\mathrm{sensor}}} = \dfrac{1}{\min_{i \in \text{A}} l_{\mathrm{actor}_i}}
\end{align}
where $A$ is the set of actors in the camera's FOV.


%% file: 4_System.tex
\section{ZHUYI-BASED AV SYSTEM}
\label{sec:system} 
The Zhuyi model has many use cases. This section presents the applications of the Zhuyi model during the AV system's pre-deployment and post-deployment phases.


\subsection{Application during Pre-Deployment} 
During the AV development phase, driving scenario-based AV testing plays a critical role in developing the AV stack that is safe under as many driving conditions as possible, despite being time-consuming. When a new feature is added to the AV stack or the system is re-designed, the stack must be retested. 
These tests run the latest AV system either in simulation or in the real world using a set of driving scenarios that consider factors such as weather, road conditions, traffic, and known pre-collision scenarios.
Each test failure, i.e., a collision during a driving scenario-based test, provides feedback to the system designers to diagnose the source of the failure and bug-fix or re-design the system to clear that test in the next iteration. 

The Zhuyi model can be implemented as a modular and independent safety evaluator that is run along with or after each test that executes a driving scenario using the latest AV system. This modular evaluation of the test will provide per-camera processing rate requirements at every time-step in a tested scenario, which can also be included in the feedback to the system designers to help design a safer and more efficient AV system.

We implemented the Zhuyi model to be run after a test. For each AV tested scenario, the scenario trace is collected which includes the states of the ego and all the actors at all the time-steps. The test is typically run with a pre-defined processing rate (e.g., $FPR_0=30$).
The Zhuyi model is executed at each time-step in the scenario trace starting from the beginning until the end of the scenario. 
As we compute the tolerable latency using Equation~\ref{eq:object} for each actor at a time, the actor's location at future time-steps is known, i.e., the size of the set $T$ is one.
The per-camera frame processing rate estimation is thus obtained by using Equation~\ref{eq:sensor}. This estimation provides AV designers with information about the dynamic per-camera processing rate requirements and the minimum processing rate for each camera in each of the tested driving scenario. 
This information can help system designers identify operational design domains where a different resource allocation for different sensors can provide a safer drive. For example, scenarios where the AV is in the right most lane with no objects and occlusions on the right-side cameras can dedicate more resources to process data from front/left/rear cameras.




\subsection{Applications during Post-Deployment} 

\begin{figure}[bp]
    \centering
    \includegraphics[width=0.95\columnwidth]{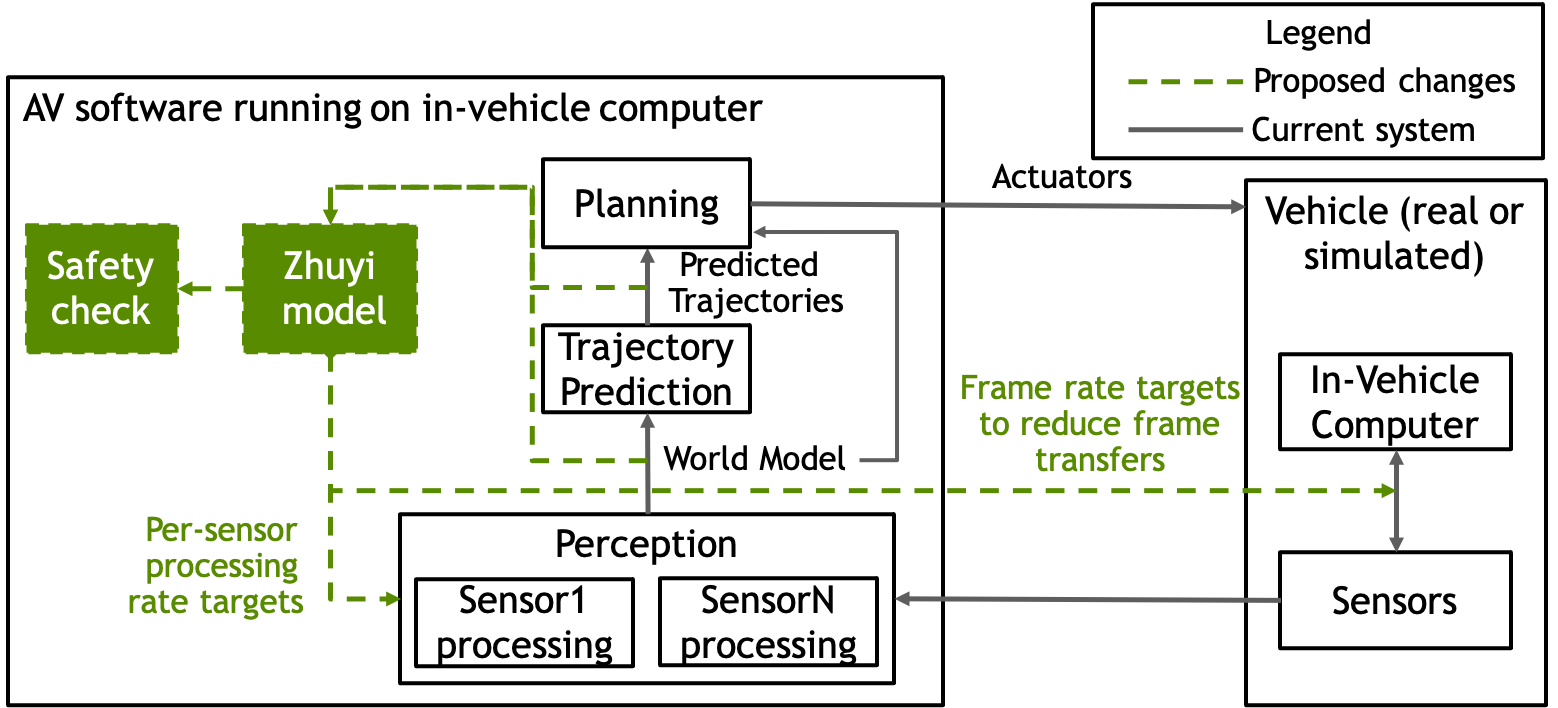}
    \caption{Zhuyi-based AV system.}
    \label{fig:zhuyi-system}
\end{figure}


The Zhuyi model can also be used online after the AV system is deployed to check that the operating frame processing rate is meeting the minimum required rate for safe operation as well as to prioritize work in times of need. 
Figure~\ref{fig:zhuyi-system} shows a typical AV system (using black boxes/lines) with the proposed functionality for safety (using green boxes/lines). The current AV system processes the per-sensor data to obtain a world model, which forms the input to the trajectory prediction module. The world model and the predicted paths (the output of the trajectory prediction) form the input to the planner. The planner sets the actuator values which are sent to the vehicle.  
The proposed change is to leverage the world model (current state) and predicted trajectories as input to execute the Zhuyi model online. The Zhuyi model provides the per-camera processing rate requirements for safety, which can then be used for safety checking and work prioritization. 

\textbf{Safety Check.} With Zhuyi's estimated per-camera requirements, the system can check whether the current per-camera processing rates are above the estimates. If not, there is a safety concern with a high potential for a collision with one or multiple surrounding actors. As a consequence, the Safety check block can send an alarm to the AV system which can take one of the following actions to maintain on-road safety:~(1)~activate an emergency back-up system, if available,~(2)~operate in a limited functionality mode that compromises non-essential tasks such as in-cockpit infotainment and user-assistance systems, or~(3)~request the system to raise the processing rate for the cameras that fall below the estimation.

\textbf{Work Prioritization.} Zhuyi's per-camera processing rate estimate can guide the work prioritization for sensors in the Perception box in Figure~\ref{fig:zhuyi-system}. The higher the processing rate estimate, the more important the images from the camera. When the current processing rates are sufficient for safety, the system can be optimized for comfort. Instead of processing each camera's images at the same frequency, the AV system could process these images at rates proportional to the estimated rates. 
If an alarm is triggered by the Safety check module, the Perception module can raise the processing rates for the cameras that are failing to reach the demand by processing fewer frames for the cameras that are operating at a rate higher than their respective estimates. 
Based on the per-camera processing rate estimate, fewer images from the sensors can be transferred to the in-vehicle computer, reducing wasted transfers of the images that the AV software does not intend to process. 

Zhuyi's per-actor tolerable latency estimates can be used to prioritize objects in the scene. The inverse of the per-actor tolerable latency estimate is proportional to the actor's importance (the higher the latency estimate, the less important the object is).
This per-actor prioritization can be used to optimize tasks that perform work for each of the actors in the scene, by truncating work for less important objects, for example. 

When extended to account for perception uncertainty, Zhuyi can be used to determine the necessary accuracy for the perception stack. As DNN models naturally present accuracy versus computation demand trade-offs (through quantization and pruning), Zhuyi can inform when to trade-off accuracy for computation reduction.

%% file: 5_Experiments.tex
\section{EXPERIMENTAL RESULTS}
\label{sec:experiment}

\subsection{Experimental Setup}

\input{Table/Scenes}

\begin{figure*}[t]
\centering
    \subfloat[Scene. \label{fig:cut-out scene}]{\includegraphics[width=0.10\linewidth]{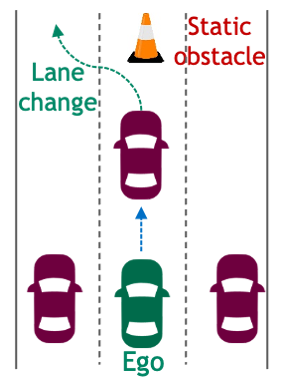}}
    \subfloat[Left camera. \label{fig:cut-out left}]{\includegraphics[width=0.21\linewidth]{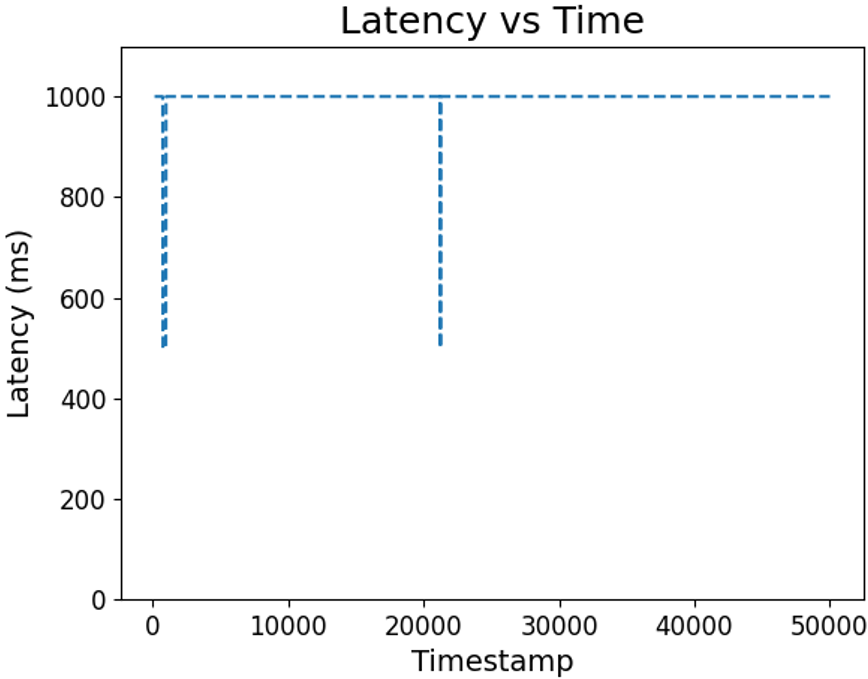}} 
    \subfloat[Front camera. \label{fig:cut-out front}]{\includegraphics[width=0.21\linewidth]{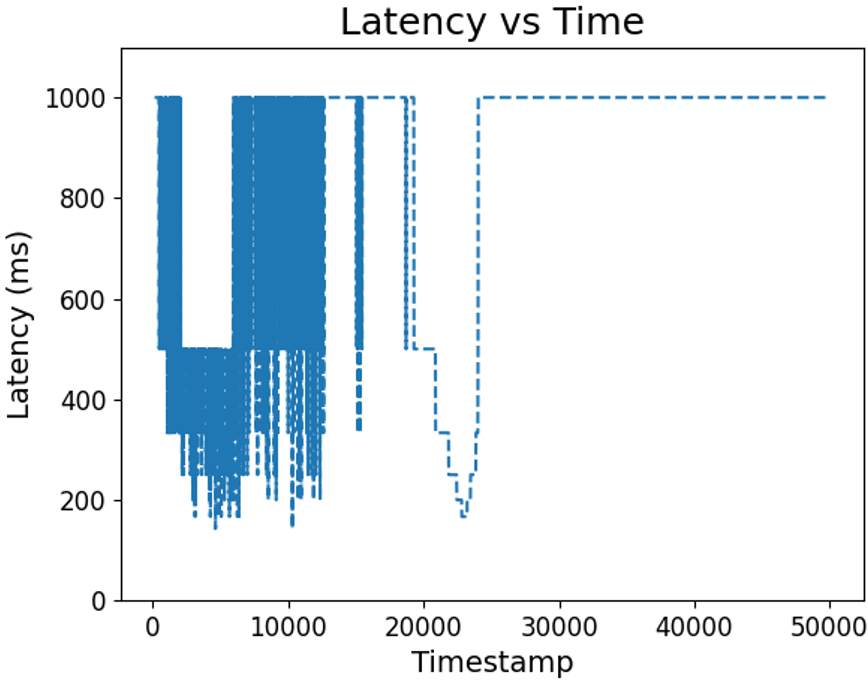}}
    \subfloat[Right camera. \label{fig:cut-out right}]{\includegraphics[width=0.21\linewidth]{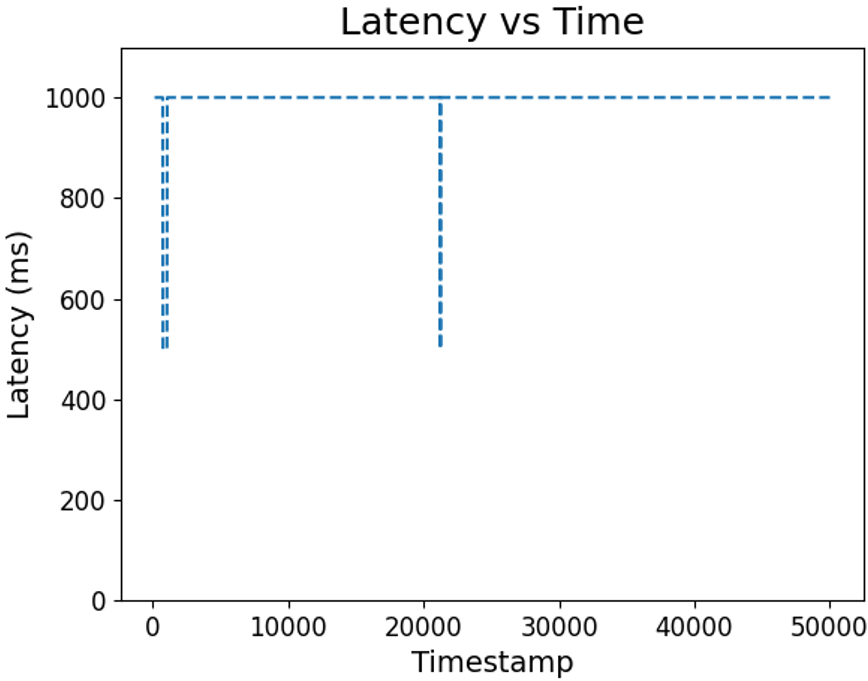}}
    \subfloat[Acceleration. \label{fig:cut-out acc}]{\includegraphics[width=0.21\linewidth]{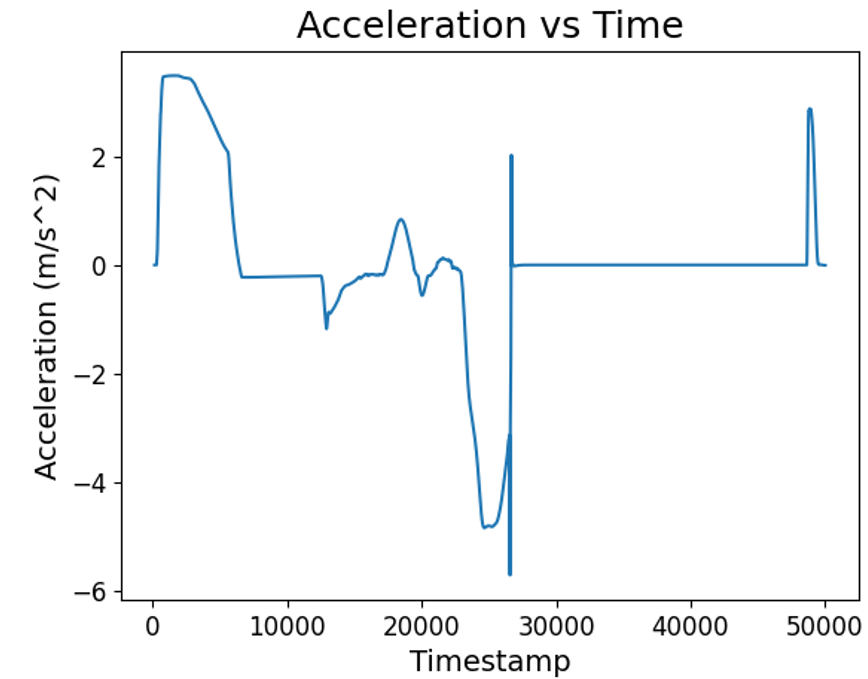}} 
    \caption[]{Per-camera latency estimates for Cut-out fast.}
    \vspace{-13pt}
    \label{fig:cut-out}
\end{figure*}

\textbf{Scenarios.} 
To evaluate the Zhuyi model and a Zhuyi-based system, we create nine driving scenarios as listed in Table~\ref{tab:scene}. 
These scenarios take place on a 3-lane road. The Cut-out scenario includes an actor in the front that cuts out of the ego's lane and reveals a static obstacle in the FOV of the ego's front camera as shown in Figure~\ref{fig:cut-out scene}. Two other actors are on the adjacent lanes, one on each side of the ego, which leaves hard braking as the only option for the ego in the center lane. The Cut-out fast scenario is the same as Cut-out except that the ego is traveling at a higher speed. 

\begin{figure*}[t]
\centering
    \subfloat[Scene. \label{fig:cut-in curve scene}]{\includegraphics[width=0.10\linewidth]{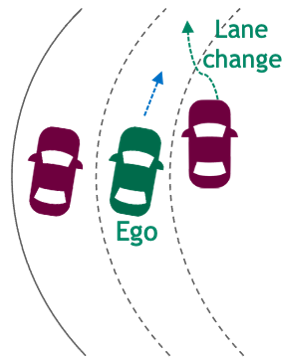}}
    \subfloat[Left camera. \label{fig:cut-in curve left}]{\includegraphics[width=0.21\linewidth]{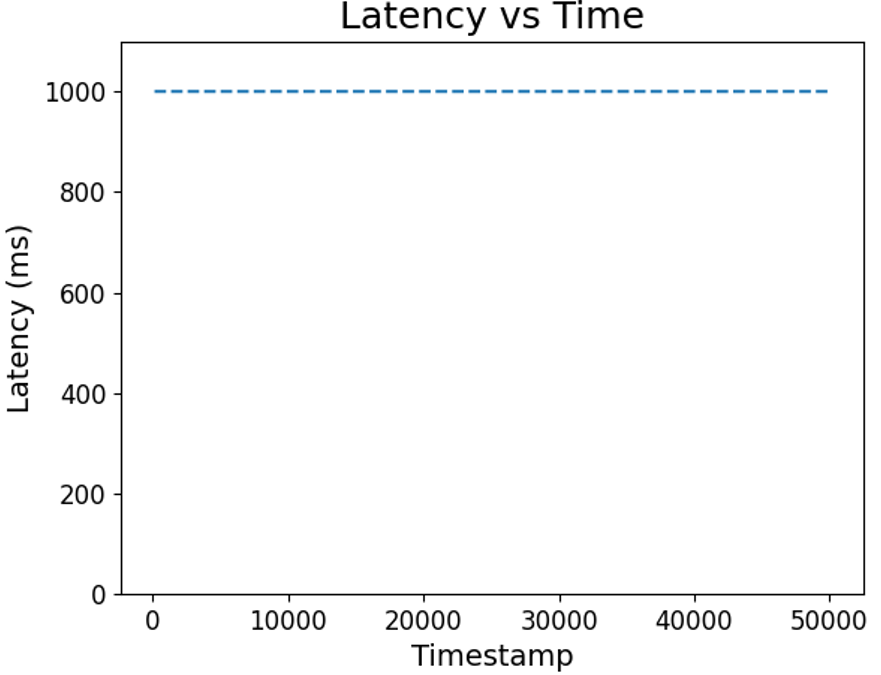}} 
    \subfloat[Front camera. \label{fig:cut-in curve front}]{\includegraphics[width=0.21\linewidth]{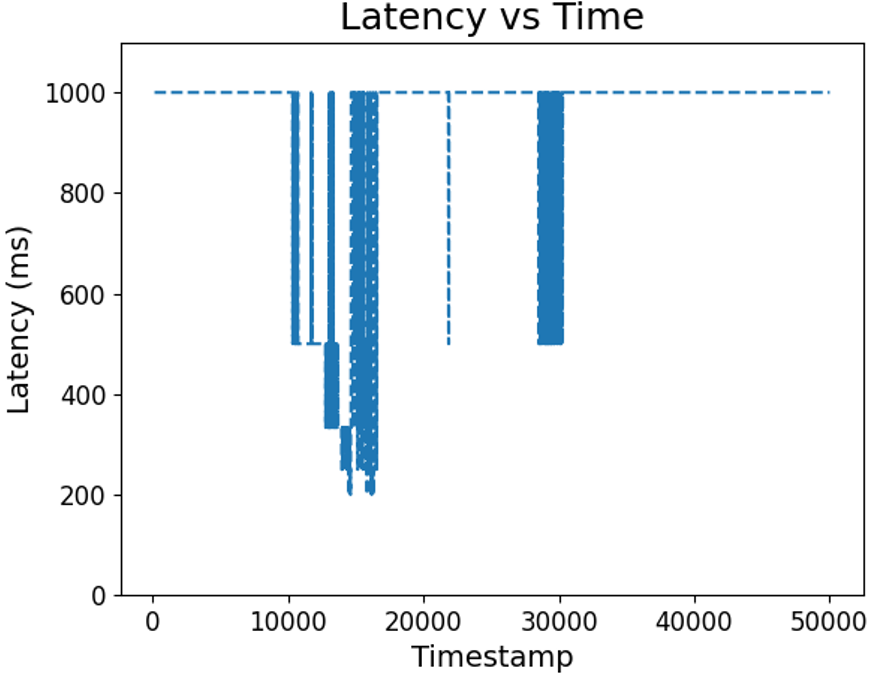}}
    \subfloat[Right camera. \label{fig:cut-in curve right}]{\includegraphics[width=0.21\linewidth]{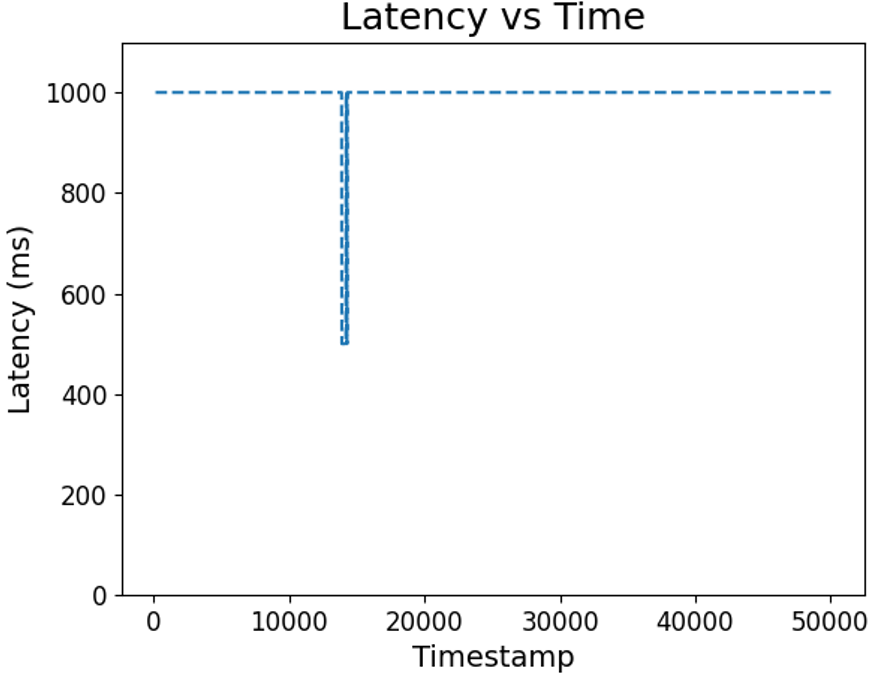}}
    \subfloat[Acceleration. \label{fig:cut-in curve acc}]{\includegraphics[width=0.21\linewidth]{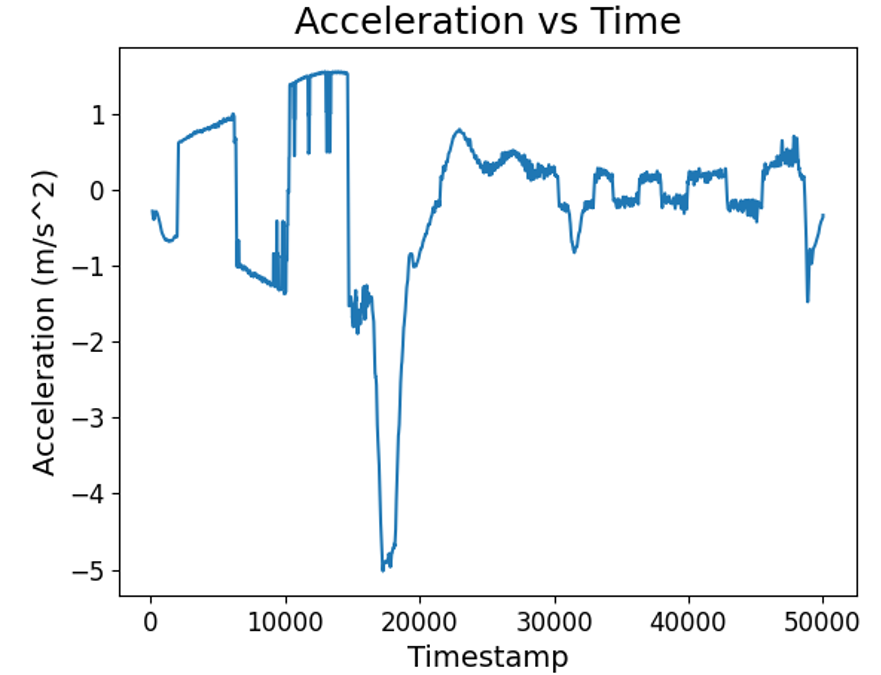}} 
    \caption[]{Per-camera latency estimates for Challenging cut-in on a curved road.}
    \vspace{-13pt}
    \label{fig:cut-in curved}
\end{figure*}

\begin{figure*}[t]
\centering
    \subfloat[Scene. \label{fig:cut-in scene}]{\includegraphics[width=0.10\linewidth]{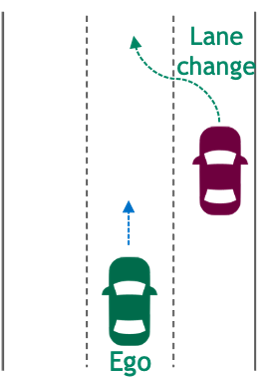}} 
    \subfloat[Left camera. \label{fig:cut-in left}]{\includegraphics[width=0.21\linewidth]{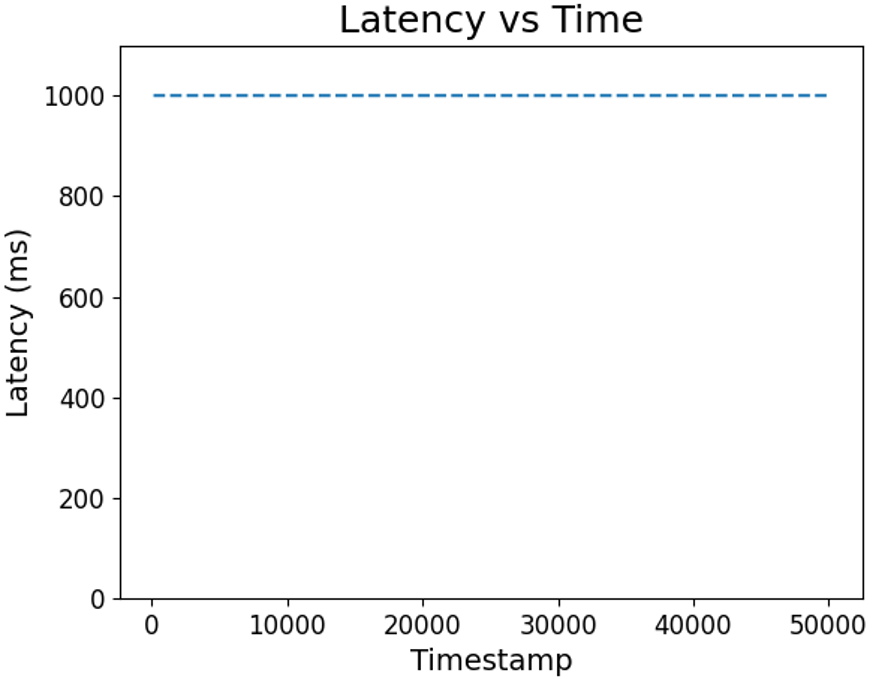}} 
    \subfloat[Front camera. \label{fig:cut-in front}]{\includegraphics[width=0.21\linewidth]{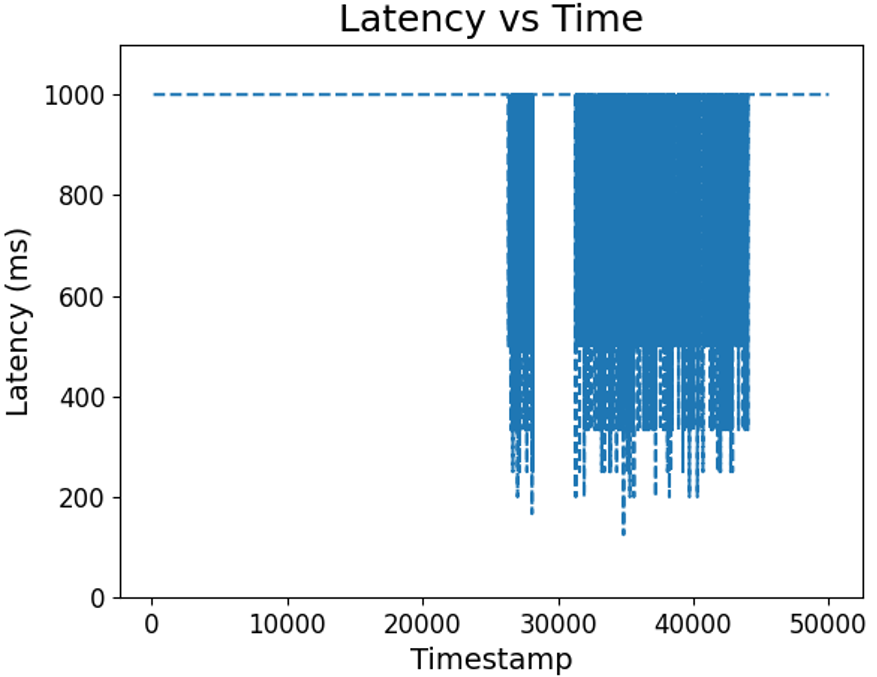}}
    \subfloat[Right camera. \label{fig:cut-in right}]{\includegraphics[width=0.21\linewidth]{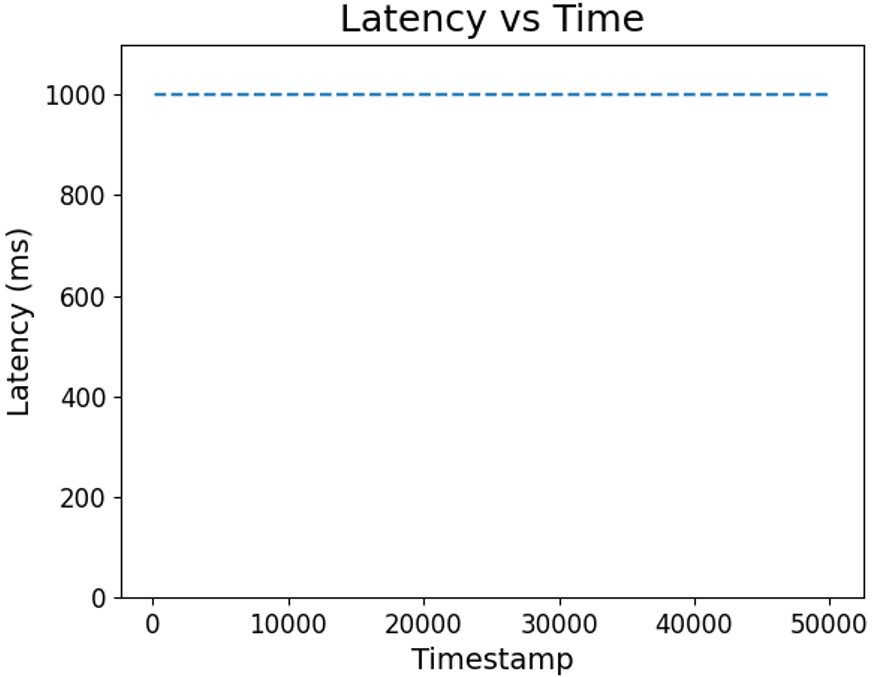}}
    \subfloat[Acceleration. \label{fig:cut-in acc}]{\includegraphics[width=0.21\linewidth]{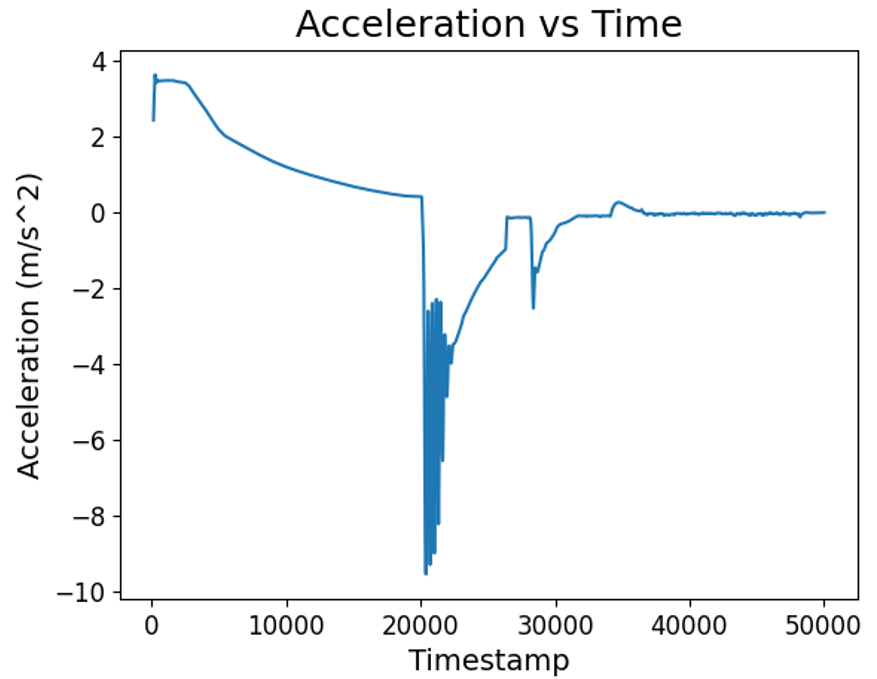}} 
    \caption[]{Per-camera latency estimates for Cut-in. }
    \vspace{-17pt}
    \label{fig:cut-in}
\end{figure*}

The three cut-in scenarios include an actor cutting in front of the ego. The Cut-in scenario is shown in Figure~\ref{fig:cut-in scene}. For the challenging cut-in scenario, the actor cuts in much closer to the ego than in the Cut-in scenario. Another actor is placed in the left lane, leaving braking as the only option. The challenging cut-in on a curved road adopts a similar setting as the challenging cut-in, but on a curved road as shown in Figure~\ref{fig:cut-in curve scene}. In the Vehicle following scenario, the ego follows the front actor at a distance of 50 meters on a highway; the actor applies sudden braking, reducing its speed to zero. 

The Front \& right activity 1, 2, and 3 scenarios target activity on only the right or left side in addition to the front side. In the first version, the ego is spawned in the left lane. An actor is launched on the right most lane, which changes lanes to the ego's adjacent lane. Another actor is launched at the back of the ego and changes lanes to the right. For the second version, the ego is spawned in the middle lane. An actor is launched in front of the ego, which cuts out to the right most lane and matches its position side to side to the ego with similar speed. Another actor is launched at the back of the ego and follows the ego. For the third version, the ego is spawned in the middle lane. An actor is launched on the right most lane, which cuts into the ego’s lane ahead of the ego.

\textbf{AV system.} We adopt the state-of-the-art NVIDIA AV stack with the NVIDIA DriveSim for environment simulation~\cite{drive,drivesim}. The AV is equipped with 5 cameras, including two front cameras (with 60 and 120 degrees FOV), two side cameras, and a rear camera. We analyze data for the front camera with 120 FOV and the two side cameras. For experimentation, we use a machine with four GPUs -- three Ampere-based A6000 GPUs for DriveSim and a Volta-based V100 GPU to execute the AV system. The machine has an AMD Epyc 7402P processor with 256GB of host DRAM.

\textbf{Zhuyi model parameters.} For the Zhuyi model described in Section~\ref{sec:methodology}, we use $C1=C2=0.9$, $C3=4.9m/s^2$, $C4=1.1$, $K=5$, $M=10$, and $L=1s/0.33s=30$, which include margins for conservative estimation.


\subsection{Validation of the Zhuyi model}


\textbf{Pre-deployment.} To validate Zhuyi's estimates during pre-deployment, we collect the ground-truth states of ego and actors at all timestamps and obtain FPR estimates for safety by running the Zhuyi model offline. As our framework only allows the same FPR  settings for all the cameras in one experiment, we validate the Zhuyi model by running the AV system with different FPR (ranging from 1 to 30) and check whether the estimated FPR for a scenario is above the \textit{m}inimum \textit{r}equired \textit{F}PR (MRF)\@. 
The MRF is the FPR above which no collision was detected in the scenario. 
Table~\ref{tab:scene} shows these values for different scenarios.
Table~\ref{tab:scene} also shows the FPR estimates provided by the Zhuyi model from experiments run at different (fixed) FPRs.
We obtain the highest estimated value across all cameras at all times for each run. Since simulations can be non-deterministic, we run a scenario with a fixed FPR ten times and show an average.
Results show that estimated FPR values are higher than MRF for all the scenarios, implying that Zhuyi provides safe estimates (we show N/A for experiments that are run with fixed FPR$\le$MRF).
We also sum the FPR estimates from the three cameras at each timestep during an experiment and show the maximum of this summation (total computation demand at a given time) across all the runs in the column named max(F$_{c1}$+F$_{c2}$+F$_{c3}$), where F$_{ci}$ refers to the FPR estimate for camera $i$. In the last column, we show the fraction of the resources offered by the system that is sufficient to maintain safe operations (assuming the system is provisioned for 30 FPR). The results show only 36\% of the resources originally provisioned for perception are required at any point in all our experiments. For some scenarios, only 3\% of the resources are needed. The extra resources can be provisioned to improve comfort or provide additional functionally.




We show the per-camera FPR estimates for three scenarios -- Cut-out fast,  Challenging cut-in on a curved road, and Cut-in in Figures~\ref{fig:cut-out},~\ref{fig:cut-in curved}, and ~\ref{fig:cut-in}, respectively. 
For the Cut-out fast scenario, the front camera processing requires 167~\textit{ms} in some time-steps as most of the activity is in the front camera's FOV. the tolerable latency for side cameras is a $\ge$500~\textit{ms} because the side activity is minimal. 
A similar trend is found for the other two scenarios (Figures~\ref{fig:cut-in curved} and~\ref{fig:cut-in}). 
For the Challenging cut-in on a curved road, the side cameras require only a maximum of 2 FPR even if an actor cuts in from an adjacent lane. 
For Cut-in, the tolerable latency for side cameras is 1000~\textit{ms} as there are no actors on the sides. This shows that the ability to prioritize per-camera's work based on the Zhuyi's FPR estimates can allow AVs to maintain safety by using a fraction of the resources when the system resources are constrained.



Across all three scenarios, we observe a strong correlation between the front camera FPR requirements and ego deceleration. 
For Cut-out fast scenario, because an actor is always in front of the ego, acceleration or deceleration could lead to a higher FPR requirement. Acceleration causes larger $d_{e1}$ and deceleration implies the following distance is too small. 
For Challenging cut-in on a curved road, the actor cuts in front of the ego not too far from it, which causes the ego to brake hard and require the highest FPR. 
For Cut-in, the highest FPR estimation is at around 28000 timestamp when the ego has the second dip of deceleration as shown in Figure~\ref{fig:cut-in acc}.
The largest deceleration in Figure~\ref{fig:cut-in acc} does not lead to the lowest latency estimate since the actor cuts in a distance away from the ego, leaving a large tolerable distance to react to the actor. However, at the second deceleration dip, the ego gets closer to the actor and brakes to meet the front actor's slow velocity, which causes a moderate FPR requirement.

\begin{figure}[t]
     \centering
     \includegraphics[width=0.5\columnwidth]{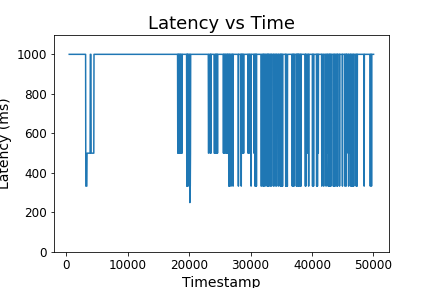}
    \caption{Post-deployment latency estimate for Cut-in.}
  \label{fig:post-cut-in}
  \vspace{-15pt}
\end{figure}

\textbf{Post-deployment.} To validate the Zhuyi model in a post-deployment setting, we implemented it in the state-of-the-art NVIDIA AV stack. The ego and actors current states are obtained from the perceived world model, and future states are obtained from predicted trajectories. Figure~\ref{fig:post-cut-in} shows the latency estimates for the front camera from the Cut-in scenario. The variance in the estimates (compared to Figure~\ref{fig:cut-in front}) are due to the differences in the current state and future predictions. A deeper analysis suggests that the differences in the current state contribute minimally to the variance, implying that the main latency differences are due to the differences in future predictions. The graphs show that the ego estimates a higher FPR requirement at the time when the other actor initiates the cut in, but the FPR requirements are lower later (compared to Figure~\ref{fig:cut-in front}), which is likely due to the changes in the actor's braking behavior and ego's driving policies. Nonetheless, the estimates are low-enough for safe operations, which suggests that the ego can process the input frames at a sufficient rate for safe operation. 

The results so far empirically validate the Zhuyi model. They show that the estimated FPR is close to and above MRF, and the FPR estimates over time per camera match the expectation based on the ego's acceleration and surrounding actor's behaviors.

\textbf{Compute demand.} The work done by the Zhuyi model is equal to $|A| \times |T| \times M \times L \times C$, where $|A|$ and $|T|$ are the number of actors and predicted trajectories per actor, respectively, and $C$ is the number of ops per iteration, which is about $100$. Beside the sequential search for $M$ steps, other aspects are parallelizable. For a scenario with 2 actors and a single future prediction, the compute demand is capped at $60$ kilo-ops. For processors offering 10+ GOPS, the Zhuyi model should execute within 2ms. 

\subsection{Model Sensitivity Analysis}

\begin{figure}[t]
    \centering
    \subfloat[$s_n=30m (98ft)$. \label{fig:sensitivity-30}]{\includegraphics[width=0.5\columnwidth]{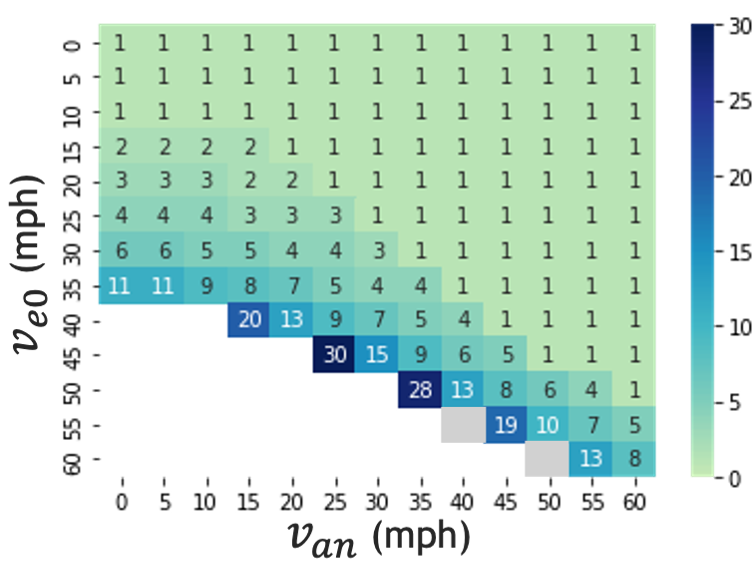}} 
    \subfloat[$s_n=100m (328ft)$. \label{fig:sensitivity-100}]{\includegraphics[width=0.5\columnwidth]{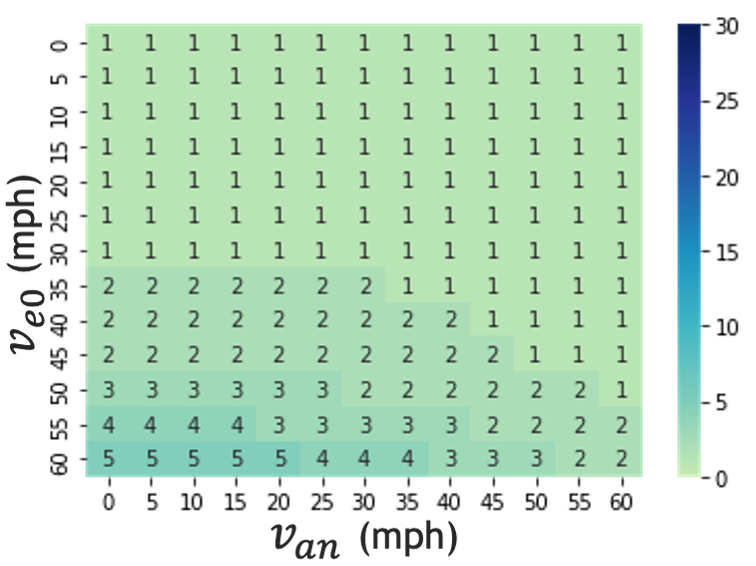}}
    \caption{Estimated minimum FPR with a fixed tolerable distance. 30$+$ FPR is shown in gray. Unavoidable collision is shown in white.}
    \label{fig:sensitivity}
    \vspace{-3pt}
\end{figure}

We analyze the relationship between velocity and estimated FPR obtained by Zhuyi. We sweep $v_{e0}$ and $v_{an}$ by fixing $s_n$, the distance the ego can travel between time $t_0$ and $t_n$ and not collide with the actor in the same lane. Results for $s_n=30m$ and $100m$ are shown in Figure~\ref{fig:sensitivity}. For an ego operating on streets (0-25mph), both Figure~\ref{fig:sensitivity-30} and Figure~\ref{fig:sensitivity-100} show that FPR$\le2$ is enough for safety.
For the ego on expressways and highways (25+ mph), a maximum of only 5 FPR is sufficient for safe operation for $s_n=100m$. For $s_n=30m$ and ego's speed over 25 mph, the FPR requirement can be high, depending on the actor's end velocity ($v_{an}$). 

High $v_{e0}$ and low $v_{an}$ combinations lead to high FPR requirements. As many such combinations are impossible (i.e., no FPR is sufficient for safe operation), only a few such combinations require high FPR (FPR$\ge$15). Even when one camera requires high FPR, other cameras may require much less FPR (as shown in Figures~\ref{fig:cut-out}-\ref{fig:cut-in}). These results reinforce that Zhuyi-based work prioritization can enable safe operation whenever the resources are constrained. 

%% file: Table/Scenes.tex
\renewcommand{\arraystretch}{1.2}

\begin{table*}[bp]
\centering
\vspace{-10pt}
\caption{Diverse driving scenarios for validation.}
\resizebox{1.0\linewidth}{!}{
\begin{tabular}{|c||c|c|c|c|c||c|c|c|c|c|c|c|c|c|c|c|c|c|c|}
\hline
 & \multicolumn{5}{c||}{\textbf{Scenario Description}}  & \multicolumn{14}{c|}{\textbf{Results}} \\ \cline{2-20}

\textbf{} & 
\textbf{Ego speed} & \textbf{Front} & \textbf{Right} & \textbf{Left} & \textbf{Min Required} &  
\multicolumn{12}{c|}{\textbf{Maximum estimated FPR when running the AV at different FPR}}                                                             
  & \multirow{2}{*}{\textbf{max}} & \multirow{2}{*}{\textbf{Fraction}}            \\ \cline{7-18}

\textbf{} & \textbf{(mph)} & \textbf{activity} & \textbf{activity} & \textbf{activity} & \textbf{FPR (MRF)} & 
\textbf{1} & \textbf{2} & \textbf{3} & \textbf{4} & \textbf{5} & \textbf{6} & \textbf{7} & \textbf{8} & \textbf{9} & \textbf{10} & \textbf{15} & \textbf{30}
 &(F$_{c1}$+F$_{c2}$+F$_{c3}$) &\\ \hline \hline

\textbf{Cut-out} & 20 & Yes & Yes & Yes & 2 &
\multicolumn{1}{c|}{N/A}        & \multicolumn{1}{c|}{4}          & \multicolumn{1}{c|}{4.8}        & \multicolumn{1}{c|}{4.8}        & \multicolumn{1}{c|}{4.8}        & \multicolumn{1}{c|}{5.8}        & \multicolumn{1}{c|}{5.4}        & \multicolumn{1}{c|}{5.4}        & \multicolumn{1}{c|}{6.1}        & \multicolumn{1}{c|}{6.4}         & \multicolumn{1}{c|}{6}           & 7           &  11    & 0.12                \\ \hline

\textbf{Cut-out fast} & 40 & Yes & Yes & Yes & 6 &
\multicolumn{1}{c|}{N/A}        & \multicolumn{1}{c|}{N/A}        & \multicolumn{1}{c|}{N/A}        & \multicolumn{1}{c|}{N/A}        & \multicolumn{1}{c|}{N/A}        & \multicolumn{1}{c|}{9.5}        & \multicolumn{1}{c|}{11.7}       & \multicolumn{1}{c|}{7.3}        & \multicolumn{1}{c|}{6.5}        & \multicolumn{1}{c|}{7.2}         & \multicolumn{1}{c|}{8}           & 7           & 32   & 0.36               \\ \hline

\textbf{Cut-in} & 70 & Yes & No & No & $<$1 &
\multicolumn{1}{c|}{-}       & \multicolumn{1}{c|}{3.9}        & \multicolumn{1}{c|}{5}          & \multicolumn{1}{c|}{5}          & \multicolumn{1}{c|}{5.9}        & \multicolumn{1}{c|}{6.1}        & \multicolumn{1}{c|}{6.5}        & \multicolumn{1}{c|}{5.9}        & \multicolumn{1}{c|}{6}          & \multicolumn{1}{c|}{6.7}         & \multicolumn{1}{c|}{7.2}         & 7.3         & 11  & 0.12                 \\ \hline

\textbf{Challenging cut-in} & 60 & Yes & Yes & No & 3 &
\multicolumn{1}{c|}{N/A}        & \multicolumn{1}{c|}{N/A}        & \multicolumn{1}{c|}{6.9}        & \multicolumn{1}{c|}{4.5}        & \multicolumn{1}{c|}{4.5}        & \multicolumn{1}{c|}{4.7}        & \multicolumn{1}{c|}{4.7}        & \multicolumn{1}{c|}{4.4}        & \multicolumn{1}{c|}{4.4}        & \multicolumn{1}{c|}{5.3}         & \multicolumn{1}{c|}{5.1}         & 5.3         & 9  & 0.10                  \\ \hline

\textbf{Challenging cut-in on a curved road} & 40 & Yes & Yes & Yes & 3 &
\multicolumn{1}{c|}{N/A}        & \multicolumn{1}{c|}{N/A}        & \multicolumn{1}{c|}{7.1}        & \multicolumn{1}{c|}{4.6}        & \multicolumn{1}{c|}{5.1}        & \multicolumn{1}{c|}{5.4}        & \multicolumn{1}{c|}{5.3}        & \multicolumn{1}{c|}{5.4}        & \multicolumn{1}{c|}{5.9}        & \multicolumn{1}{c|}{5.4}         & \multicolumn{1}{c|}{6.3}         & 5.7         & 11   & 0.12              \\ \hline

\textbf{Vehicle following} & 70 & Yes & No & No & $<$1 & 
\multicolumn{1}{c|}{2}          & \multicolumn{1}{c|}{3}          & \multicolumn{1}{c|}{3.1}        & \multicolumn{1}{c|}{3.1}        & \multicolumn{1}{c|}{4}          & \multicolumn{1}{c|}{6.6}        & \multicolumn{1}{c|}{4.8}        & \multicolumn{1}{c|}{7.5}        & \multicolumn{1}{c|}{4.9}        & \multicolumn{1}{c|}{4.9}         & \multicolumn{1}{c|}{6}           & 8           & 32    & 0.36             \\ \hline

\textbf{Front \& right activity 1} & 40 & Yes & Yes & No & $<$1 & 
\multicolumn{1}{c|}{1}          & \multicolumn{1}{c|}{1}          & \multicolumn{1}{c|}{1}          & \multicolumn{1}{c|}{1}          & \multicolumn{1}{c|}{1}          & \multicolumn{1}{c|}{1}          & \multicolumn{1}{c|}{1}          & \multicolumn{1}{c|}{1}          & \multicolumn{1}{c|}{1}          & \multicolumn{1}{c|}{1}           & \multicolumn{1}{c|}{1}           & 1           & 3   & 0.03                \\ \hline

\textbf{Front \& right activity 2} & 40 & Yes & Yes & No & $<$1 & 
\multicolumn{1}{c|}{2}          & \multicolumn{1}{c|}{2.4}        & \multicolumn{1}{c|}{2.2}        & \multicolumn{1}{c|}{2.3}        & \multicolumn{1}{c|}{2.7}        & \multicolumn{1}{c|}{2.9}        & \multicolumn{1}{c|}{2.9}        & \multicolumn{1}{c|}{2.8}        & \multicolumn{1}{c|}{2.7}        & \multicolumn{1}{c|}{3.1}         & \multicolumn{1}{c|}{3.4}         & 3.1         & 7     & 0.08             \\ \hline

\textbf{Front \& right activity 3} & 60 & Yes & No & Yes & $<$1 &
\multicolumn{1}{c|}{2}          & \multicolumn{1}{c|}{4}          & \multicolumn{1}{c|}{5}          & \multicolumn{1}{c|}{4.8}        & \multicolumn{1}{c|}{5.2}        & \multicolumn{1}{c|}{5.6}        & \multicolumn{1}{c|}{5.6}        & \multicolumn{1}{c|}{5.9}        & \multicolumn{1}{c|}{6}          & \multicolumn{1}{c|}{5.4}         & \multicolumn{1}{c|}{6.6}         & 5.7         & 10    & 0.11             \\ \hline
\end{tabular}
}
\label{tab:scene}
\end{table*}

%% file: 2_Related.tex
\section{RELATED AND FUTURE WORK}
\label{sec:related} 

AV safety policies such as Responsibility-Sensitive Safety (RSS)~\cite{shalev2017formal} and Safety Force Field (SFF)~\cite{nister2019safety} define the high-level rules for safety. Both RSS and SFF focus on how to make planning and control decision to avoid collision while lack of insights on the safety-aware AV system design. Safety Score~\cite{zhao2020safety} proposes the safety score to guide the safety-aware AV computing system design. Unfortunately, the safety-score only focuses on the the high-level latency of each perception module and does not provide perception requirements for safe AVs. Suraksha~\cite{suraksha} suggests that 10 FPS is enough to ensure safety for AV with one camera setting for the studied scenarios, which shows the potential for better optimization for sensor FPS setting. However, the grid search adopted in Suraksha could easily become infeasible in multi-camera setting. RoboRun~\cite{boroujerdian2021roborun} employs a model to dynamically adjust accuracy and speed of algorithms based on object density for drones. Zhuyi considers more factors (such as actor states and dynamics) and is applied to AVs.
A recent study~\cite{chin2018domain} suggests adjusting object detection accuracy based on the scenario to balance resource constraints, which can be combined with Zhuyi.
%
%
Future research directions include extending the Zhuyi model to consider perception uncertainty to facilitate trading-off perception model accuracy for performance, accounting for occlusions in the world model, and incorporating yet-to-be-detected objects.

%% file: 6_Conclusion.tex
\section{CONCLUSION}
\label{sec:conclusion}
Zhuyi quantifies the lowest safe sensor frame processing rate by considering the AV's kinematics and the surrounding actors.  A Zhuyi-based AV system provides online safety checks and workload prioritization based on the per-camera FPR estimations. Zhuyi's estimated FPRs are conservative yet only require 36\% or less of the compute resources for perception compared to the fixed 30-FPR system, without compromising safety during operation.


\vspace{-3pt}
\section{Acknowledgments} \label{sec:ack}
This work was supported by IARPA under contract number 2022-21102100013 and sponsored by the ADA (Applications Driving Architectures) Center and C-BRIC (Center for Brain-inspired Computing), two of six centers in JUMP, a Semiconductor Research Corporation (SRC) program sponsored by DARPA.

\vspace{-3pt}